\documentclass{article}
\usepackage{spconf,amsmath,graphicx}
\usepackage{xcolor}
\usepackage[normalem]{ulem}
\newcounter{ToDo}
\newcounter{gaocomm} 
\newcounter{Note}
\definecolor{blue-violet}{rgb}{0.00,0.75,0.90}
\definecolor{mygreen}{rgb}{0.0, 0.5, 0.0}
\definecolor{awesome}{rgb}{1.0, 0.13, 0.32}
\definecolor{bostonuniversityred}{rgb}{1.0, 0.0, 0.0}
\title{A Magnetic Framelet-Based Convolutional Neural Network for Directed Graphs}
\name{Lequan Lin and Junbin Gao}
\address{Discipline of Business Analytics, The University of Sydney Business School\\ The University of Sydney, Camperdown, NSW 2006, Australia\\
llin0615@uni.sydney.edu.au, junbin.gao@sydney.edu.au}
\begin{document}
%\ninept
%
\maketitle
\begin{abstract}
Spectral Graph Convolutional Networks (spectral GCNNs), a powerful tool for analyzing and processing graph data, typically apply frequency filtering via Fourier transform to obtain representations with selective information. Although research shows that spectral GCNNs can be enhanced by framelet-based filtering, the massive majority of such research only considers undirected graphs. In this paper, we introduce Framelet-MagNet, a magnetic framelet-based spectral GCNN for directed graphs (digraphs). The model applies the framelet transform to digraph signals to form a more sophisticated representation for filtering. Digraph framelets are constructed with the complex-valued magnetic Laplacian, simultaneously leading to signal processing in both real and complex domains. We empirically validate the predictive power of Framelet-MagNet over a range of state-of-the-art models in node classification, link prediction, and denoising.
\end{abstract}
\section{Introduction}
\label{sec:intro}
Recent years have witnessed the surging popularity of research on graph convolutional neural networks (GCNNs) \cite{zhang2019graph}. Through the integration of graph signals and topological structures in graph convolution, GCNNs usually produce more valuable insights than models that analyze data in isolation. Especially, spectral GCNNs define their graph convolution in the frequency domain, enabling the filtering of different frequency components in graph signals. However, the majority of studies on signal processing in spectral GCNNs only focus on undirected graphs \cite{xu2019graph, zheng2021howframelet, yang2022quasi}. In this paper, we aim to extend framelet-based signal processing to spectral GCNNs for directed graphs (digraphs).
%\GaoC{digraph is only defined in absract. Better redefine it here. ie. directed graphs (digraphs)} 

Many directional relationships are naturally modelled as digraphs, such as citation relationships \cite{an2004characterizing}, website hyperlinks \cite{abedin2009graph}, and road directions \cite{li2018diffusion}. Using graph edges to represent directional information backbones the exploration of more aspects of the underlying data, which usually provides more useful findings. Nonetheless, while spectral GCNNs assume the eigendecomposition of symmetric graph Laplacian to provide real-valued eigenvalues and orthonormal eigenvectors, the digraph Laplacian is asymmetric. Converting digraphs to undirected graphs facilitates the extension of spectral methods to digraphs, but destroys the digraph structure. Therefore, many recent studies engage in designing symmetric digraph Laplacian that can preserve the directional information \cite{monti2018motifnet, ma2019spectral, Tong2020DigraphIC, Zhang2021MagNetAN}. Magnetic Laplacian \cite{ Zhang2021MagNetAN, fanuel2017magnetic} is one of the most successful instances. It is a complex-valued Hermitian matrix, whose real part shows edge existence, and imaginary part indicates edge directions. Magnetic Laplacian-based digraph networks exploit magnetic Laplacian in classic spectral GCNN architectures and have demonstrated their power in various graph tasks \cite{Zhang2021MagNetAN}.

Classic spectral GCNNs adopt Fourier transform in their convolutional layer. Converting graphs signals to the Fourier frequency domain allows the processing of signal frequencies, but only from a global perspective. More specifically, although we can detect signal frequencies, we cannot identify their position in the graph. To investigate both global and local information, we can ensemble spectral GCNNs with framelet transform instead. The framelet frequency domain is composed of graph framelets, which are constructed through dilation and translation of a set of localized scaling functions. Nevertheless, most existing wavelet/framelet-based networks are solely applicable to undirected graphs \cite{xu2019graph, zheng2021howframelet, yang2022quasi}. Although SVD-GCN \cite{zou2022simple} accomplishes digraph framelet transform via singular value decomposition (SVD) of the asymmetric digraph Laplacian, its theoretical rationale is very vague, for example, how the Laplacian frequency can be linked to the signal frequency in the SVD domain.

In this paper, we propose Framelet-MagNet, a magnetic Laplacian-assisted framelet-based spectral GCNN for digraphs. Multiresolution Analysis enables us to construct digraph framelets with the magnetic Laplacian and a filter bank \cite{dong2017sparse, zheng2022decimated}. In addition, we also construct quasi-framelet directly in the frequency domain to impose double regulation on digraph signals \cite{yang2022quasi}. We exploit Chebyshev polynomial approximation for fast framelet transform and reconstruction.  %inversion. 

The contributions of our work are threefold. (1) To our best knowledge, this is the first attempt to construct a framelet-based digraph GCNN without discarding the role of Laplacian eigendecomposition. (2) We realize framelet-based convolution on digraph data in both real and complex domains, which enriches the basis for signal processing. (3) Vast experiment results validate the superiority of our model over the state-of-the-art approaches in various digraph tasks. 

\section{The Method}
% Our model, Framelet-MagNet, is a spectral GCNN for digraphs that adopts framelet-based convolution, in which graph signals are processed in the framelet frequency domain before being converted back to the original spatial domain. We design a tight framelet system using the magnetic Laplacian, called the Magnetic Graph Framelet System (MGFS), as the basis for framelet transform, such that no information loss will occur during the whole process. 

\subsection{Magnetic Laplacian}
Magnetic Laplacian, a complex-valued Hermitian matrix, is a digraph representation whose real part indicates edge existence and imaginary part shows edge directions. Let $\mathcal{G}_d\{\mathcal{V}, \mathcal{E}, \mathbf{A}\}$ be a digraph, where $\mathcal{V}$ is a set of $N$ vertices, $\mathcal{E}$ is a set of edges, and $\mathbf{A}$ is the adjacency matrix. The first step to construct the magnetic Laplacian is to decompose the adjacency matrix into a symmetric part $\mathbf{A}_{sym}$ and a skew-symmetric part $\mathbf{A}_{skew}$ as following:
\begin{align*}
    \mathbf{A}_{sym} (i,j) & = \frac{1}{2} (\mathbf{A} (i,j)+\mathbf{A} (j,i)), \quad 1\leq i,j \leq N,\\
    \mathbf{A}_{skew}(i,j) &= \mathbf{A}(i,j) - \mathbf{A}(j,i), \quad 1\leq i,j \leq N.
\end{align*}
The degree matrix corresponding to the symmetric part is denoted as $\mathbf{D}_{sym}$. Then, the magnetic Laplacian (normalized) can be defined as
\begin{equation}
    \mathbf{\mathcal{L}}^{(q)} = \mathbf{I} - \mathbf{\Psi}^{(q)} \odot \left(\mathbf{D}_{sym}^{-\frac{1}{2}} \mathbf{A}_{sym} \mathbf{D}_{sym}^{-\frac{1}{2}}\right),
\end{equation}
where
\begin{equation*}
    \mathbf{\Psi}^{(q)} (i,j) = \exp{(2 \pi \mathrm{i} q \mathbf{A}_{skew}(i,j))}, \quad 1\leq i,j \leq N.
\end{equation*}
Here $q$ in $\mathbf{\Psi}^{(q)}$ is a non-negative electric charge parameter \cite{Zhang2021MagNetAN, fanuel2017magnetic}. %\GaoC{Cite the original paper magnet?} 
$q$ is designed as a hyperparameter whose range is between $0$ and $0.25$, where higher $q$ allows the Laplacian to encode more directional information. When we set $q = 0$, no directional information will be stored. Since the complex component $\mathbf{\Psi}^{(q)}  \odot \mathbf{A}_{sym}$ is Hermitian, and the degree matrix is diagonal, the magnetic Laplacian is Hermitian as well. According to \cite{Zhang2021MagNetAN}, the magnetic Laplacian is positive semi-definite, hence it supports the eigendecomposition required by spectral GCNNs.

\subsection{Magnetic Graph Framelet Transform}

\begin{figure}[t]
\begin{minipage}[b]{1.0\linewidth}
  \centering
  \centerline{\includegraphics[width=9cm]{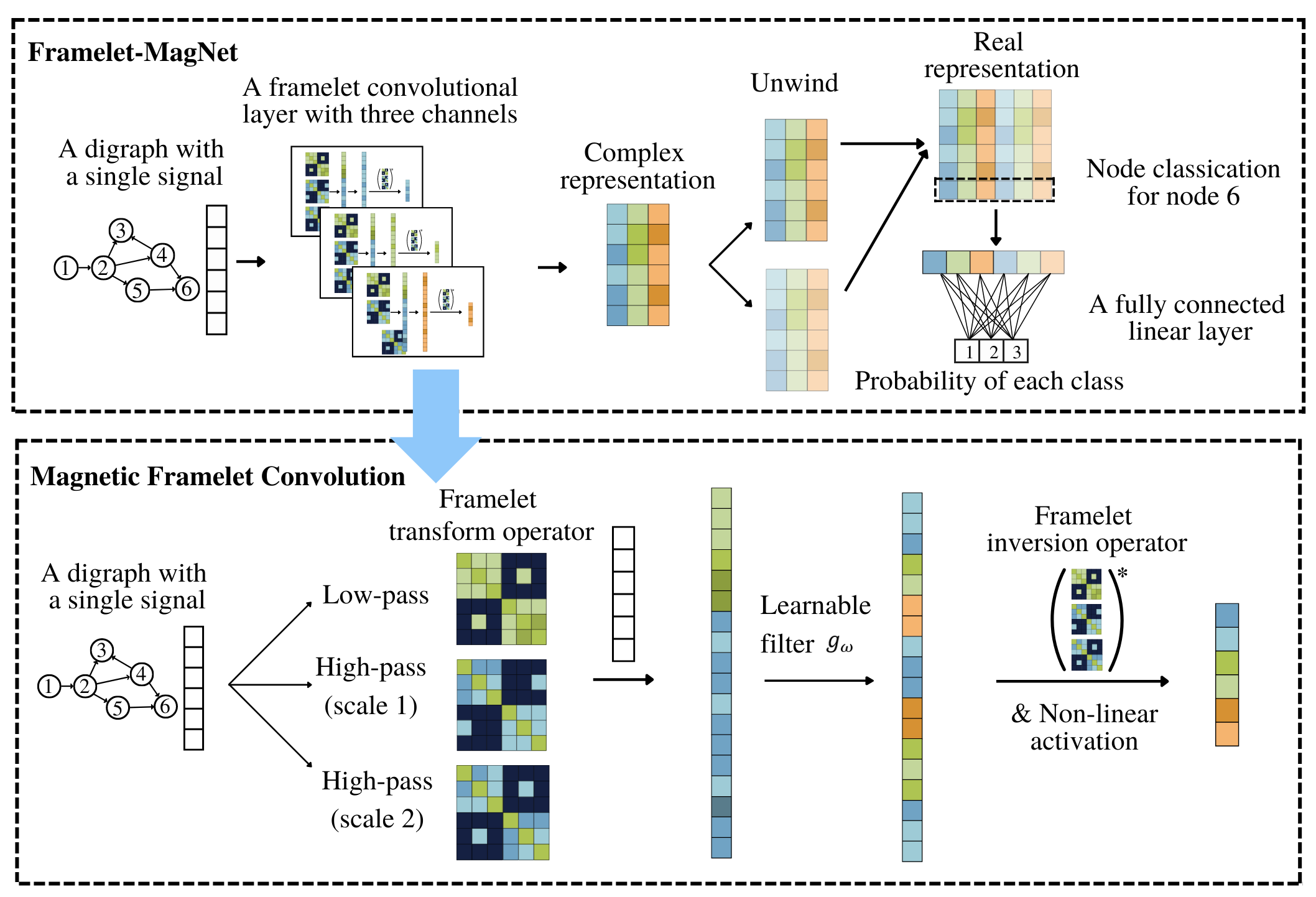}}
%  \vspace{2.0cm}
\end{minipage}
\caption{Framelet-MagNet for a node classification task. In the framelet convolutional layer, graph signal is decomposed into low-pass and high-pass components stored in a long vector, then processed by filter $g_{\omega}$ before we convert it back to the spatial domain and apply activation. Then, we unwind the new representation and apply a fully connected linear layer for classification.}
\label{fig:method}
\end{figure}

The most important component in framelet-based spectral GCNNs is the framelet convolution. We firstly transform graph signals to the framelet frequency domain for filtering, then convert the processed data back to the spatial domain with the framelet reconstruction function. Intuitively, we desire no information loss during the whole process, which means we expect the framelet transform to be ``tight''. Accordingly, we design the Magnetic Graph Framelet Transform (MGFT), which is a tight framelet transform defined on digraphs. The fundamental principle of MGFT is to incorporate magnetic Laplacian in the traditional undecimated tight framelet transform on undirected graphs. For a review of the traditional approaches and their theoretical background, we refer readers to \cite{dong2017sparse, zheng2022decimated, hammond2011wavelets}.

As the magnetic Laplacian is positive semi-definite, we can write it as $\mathcal{L}^{(q)} = \mathbf{U}\mathbf{\Lambda}\mathbf{U}^*$, where $*$ denotes conjugate transpose. We let $\{(u_k, \lambda_k)\}_{k=0}^{N-1}$ be the eigenvector and eigenvalue pairs of $\mathcal{L}^{(q)}$. With the transition position $n \in \mathcal{V}$ and the dilation level 
%\GaoC{let us use level to avoid confusion with scale. In theory the scale used here is actually 2.} %scale
$s = 1, ..., S$, we define the low-pass and high-pass magnetic graph framelets $\rho_{n,s}^{(q)}$ and $\varrho_{n,s,r}^{(q)}$ as
\begin{equation} \label{eq_magframelets}
\begin{aligned}
    \rho_{n,s}^{(q)} (m) &= \sum_{k=0}^{N-1} u_k(m) \hat{\zeta}_0 \left( \frac{\lambda_k}{2^s} \right) u_k^* (n),\\
    \varrho_{n,s,r}^{(q)} (m) &= \sum_{k=0}^{N-1} u_k(m) \hat{\zeta}_r \left( \frac{\lambda_k}{2^s}  \right) u_k^* (n), \; 1 \leq r \leq R,
\end{aligned}
\end{equation}
where $Z = \{\zeta_0,...,\zeta_R\}$ is a set of real-valued scaling functions. The low-pass and high-pass framelets will decompose graph signals into low and high frequency components during transform. According to \cite{dong2017sparse}, we can find the appropriate set of scaling functions via Multiresolution Analysis. Basically, we derive scaling functions from a filter bank $a = \{a_0,...,a_R\}$ defined in the spatial domain with the following relationship
\begin{equation}
    \hat{\zeta}_r (2\delta) = \hat{a}_r(\delta) \hat{\zeta}_0(\delta),
\end{equation}
for $r = 1,...,R$ and any $\delta \in {\rm I\!R}$. Then, ``quasi-framelet'' proposed by Yang et al. \cite{yang2022quasi} relaxes the requirement of Multiresolution Analysis by straightforwardly constructing a quasi filter bank $b = \{b_0,...,b_R\}$ in the Fourier domain. By definition, $b$ should satisfy the identity condition  
\begin{equation}
    \sum_{r=0}^R b_r(\delta)^2 \equiv 1, \quad \forall \delta \in [0, \pi],
\end{equation}
such that the value of $b_0$ decreases from 1 to 0 while the value of $b_R$ increases from 0 to 1 over the Fourier domain $[0,\pi]$. This will allow framelet convolution to impose ``double regulation'' on the graph signals. More specifically, graph signals are regulated by not only the learnable filter in traditional convolutions, but also the modulation functions $b_0,...,b_R$, where $b_0$ and $b_R$ attenuate high and low frequency components, and the rest regulates frequency in between. In the following discussions, we denote $\hat{a} = \{\hat{a}_0,..., \hat{a}_R\}$ and $b = \{b_0,...,b_R\}$ collectively as $z = \{z_0, ..., z_R\}$ for simplicity. With a single signal $x \in {\rm I\!R}^N$, we can define MGFT as
\begin{equation}
    \begin{aligned}
        \left\{{\langle x, \rho_{n,S}^{(q)} \rangle}\right\}_{n \in \mathcal{V}} &= \mathcal{F}_{0,S}^{(q)}  x,\\
        \left\{{\langle x, \varrho_{n,s,r}^{(q)} \rangle}\right\}_{n \in \mathcal{V}} &= \mathcal{F}_{r,s}^{(q)}  x, \quad 1 \leq r \leq R, 1 \leq s \leq S,
    \end{aligned}
\end{equation}
where for $s = 1$,
$\mathcal{F}_{r,1}^{(q)}  = \mathbf{U} z_r \left(\frac{\mathbf{\Lambda}}{2^{M}}\right) \mathbf{U}^* $
and for $s = 2,...,S$,
\[\mathcal{F}_{r,s}^{(q)}  = \mathbf{U} z_r \left(\frac{\mathbf{\Lambda}}{2^{M+s-1}}\right) z_0 \left(\frac{\mathbf{\Lambda}}{2^{M+s-2}}\right) \cdots z_0 \left(\frac{\mathbf{\Lambda}}{2^{M}}\right) \mathbf{U}^* .\]
In these equations, $M$ is the smallest number such that $\lambda_{\max} \leq 2^M \pi$, and $ z_r \left(\frac{\mathbf{\Lambda}}{2^{M}}\right) = \text{diag}\left( z_r \left(\frac{\mathbf{\lambda_k}}{2^{M}}\right)\right)$ for $r = 0,..., R$ and $k = 0,...,N-1$. We can use a vertically stacked transform matrix $\mathcal{F}^{(q)} = \left[\mathcal{F}_{0,S}^{(q)}; \mathcal{F}_{1,1}^{(q)}; ...; \mathcal{F}_{1,S}^{(q)};...; \mathcal{F}_{R,S}^{(q)} \right]$ to express magnetic framelete transform more concisely as
\begin{equation}
    F^{(q)} = \mathcal{F}^{(q)} x,
\end{equation}
where $F^{(q)}$ can be considered as framelet coefficients. Then, magnetic framelet reconstruction is given by
\begin{equation}
    \Tilde{x} = \mathcal{F}^{(q)^*} F^{(q)}.
\end{equation}
Recall that we expect MGFT to be tight, that is, $x = \Tilde{x}$. We can achieve this by selecting appropriate filter banks. Existing examples include \texttt{Haar} \cite{dong2017sparse},  \texttt{Linear} \cite{dong2017sparse},  \texttt{Quadratic} \cite{dong2017sparse},  \texttt{Sigmoid} \cite{yang2022quasi}, and  \texttt{Entropy} \cite{yang2022quasi}. We will exploit them in our experiment.

Employing the fast computation proposed in \cite{hammond2011wavelets}, we approximate the filter bank $z$ with with Chebyshev polynomials $\Tilde{T}_r(\cdot)$, $r = 0,...,R$. We propose Fast Magnetic Framelet Transform (FMFT) that defines the transform operator as $\Tilde{\mathcal{F}}^{(q)} =\left[\Tilde{\mathcal{F}}_{0,S}^{(q)}; \Tilde{\mathcal{F}}_{1,1}^{(q)}; ...; \Tilde{\mathcal{F}}_{1,S}^{(q)};...; \Tilde{\mathcal{F}}_{R,S}^{(q)} \right]$, where for s = 1, $\Tilde{\mathcal{F}}_{r,1}^{(q)} = \mathcal{T}_r (2^{-M} \mathcal{L}^{(q)})$, and for $s = 2,...,S$, $\mathcal{F}_{r,s}^{(q)} = \mathcal{T}_r (2^{-M-s+1} \mathcal{L}^{(q)}) \mathcal{T}_0 (2^{-M-s+2} \mathcal{L}^{(q)}) \cdots  \mathcal{T}_0 (2^{-M} \mathcal{L}^{(q)})$. Accordingly, the framelet transform and reconstruction are approximated by $F^{(q)} \approx \Tilde{\mathcal{F}}^{(q)} x$ and $x \approx \Tilde{\mathcal{F}}^{(q)^*} F^{(q)}$. 

\subsection{Framelet-Magnet}
With the FMFT, we define the $i^{th}$ magnetic framelet convolutional layer as
\begin{equation*}
    \sigma (g_{\omega_i} \ast X_{i-1}) = \sigma\!\left(\Tilde{\mathcal{F}}^{(q)^*}\!\left(\text{diag}(\omega_i)\!\left(\!\Tilde{\mathcal{F}}^{(q)} (\mathbf{X}_{i-1} \mathbf{W}_i) \right)\!\right)\!\right), 
\end{equation*}
where $\sigma$ is a non-linear activation function, $g_{\omega_i} = \text{diag}(\omega_i)$ is a learnable filter, $\mathbf{X}_{i-1}$ is an $N \times D_{i-1}$ feature matrix with $\mathbf{X}_0$ being the original graph feature matrix, and $\mathbf{W}_i$ is a $D_{i-1} \times D_i$ matrix, where $D_{i-1}$ and $D_i$ are dimensions of the input and output channels. 

Framelet-Magnet is composed of one or multiple magnetic framelet convolutional layers, an unwind operator, and a fully connected linear layer before the output layer. Suppose we have $C$ convolutional layers, then we will obtain an $N \times D_C$ complex-valued graph representation. The purpose of the unwind operator is to unwind this representation to an $N \times 2D_C$ real-valued representation for further processing. In node classification tasks, we use this real representation directly for prediction as in \textbf{Fig. \ref{fig:method}}. In link prediction tasks, on the other hand, we will concatenate the rows corresponding to the node pair connected by each edge before feeding it to the following layers. 

\section{Experiments}
\label{sec:pagestyle}

\begin{table*}[t!]
\footnotesize
\caption{Experiment Results: Node Classification and Link Prediction Accuracy (\%)}
    \label{Table_results}
    \centering
    \setlength{\tabcolsep}{0.8mm}{
    \begin{tabular}{c|c|c|c|c|c|c|c|c|c}\hline
          & \multicolumn{3}{c}{\textbf{Node Classification}} \vline & \multicolumn{3}{c}{\textbf{Link Existence}} \vline & \multicolumn{3}{c}{\textbf{Link Direction}}\\
         \hline
          \textbf{Models} & \textbf{CORA\_ML} & \textbf{CITESEER} &  \textbf{CORNELL} & \textbf{CORA\_ML} & \textbf{CORNELL} & \textbf{CHAMELEON} & \textbf{CORA\_ML} & \textbf{CORNELL} & \textbf{CHAMELEON}\\
        \hline
        ChebNet \cite{defferrard2016convolutional}
        & 60.8  $\pm$ 3.3 & 53.3 $\pm$ 2.6 & 74.1 $\pm$ 2.8 & 50.1 $\pm$ 0.1 & 49.7 $\pm$ 1.4 & 50.1 $\pm$ 0.0 & 50.1 $\pm$ 0.2 & 49.6 $\pm$ 10.3 & 50.0 $\pm$ 0.0 \\
        
        GCN \cite{kipf2017semisupervised}
        & 69.7 $\pm$ 2.0 & 60.1 $\pm$ 2.6 & 42.4 $\pm$ 5.7 & 73.1 $\pm$ 5.3 & 51.1 $\pm$ 3.7 & \textbf{89.8 $\pm$ 0.5} & 79.1 $\pm$ 1.5 & 52.0 $\pm$ 2.8 & 96.8 $\pm$ 0.6\\
        
        APPNP \cite{klicpera2019predict}
        & 79.4 $\pm$ 2.7 & 66.7 $\pm$ 2.0 & 42.7 $\pm$ 5.5 & 69.5 $\pm$ 3.9 & 61.4 $\pm$ 8.0 & 87.1 $\pm$ 4.9 & 81.9 $\pm$ 0.9 & 70.3 $\pm$ 10.9 & 97.4 $\pm$ 0.2\\
        
        GraphSAGE \cite{hamilton2017inductive}
        & 78.7 $\pm$ 1.1 & 66.4 $\pm$ 1.3 & 69.2 $\pm$ 3.5 & 67.2 $\pm$ 3.7 & 63.5 $\pm$ 9.4 & 86.0 $\pm$ 0.5 & 69.1 $\pm$ 0.5 & 69.0 $\pm$ 7.4 & 94.2 $\pm$ 0.3\\
        
        GIN \cite{xu2019how}
        & 78.7 $\pm$ 1.8 & 63.9 $\pm$ 2.2 & 48.1 $\pm$ 5.0 & 75.0 $\pm$ 3.4 & 65.5 $\pm$ 8.5 & 83.9 $\pm$ 7.1 & 84.2 $\pm$ 0.9 & 77.0 $\pm$ 7.1 & 97.6 $\pm$ 0.2\\
        
        GAT \cite{velivckovic2017graph}
        & 81.2 $\pm$ 2.0 & 66.2 $\pm$ 1.7 & 45.4 $\pm$ 10.4 & 50.0 $\pm$ 0.2 & 51.4 $\pm$ 3.5 & 50.4 $\pm$ 1.0 & 50.0 $\pm$ 0.6 & 50.7 $\pm$ 3.1 & 51.6 $\pm$ 2.2\\
        
        DGCN \cite{tong2020directed}
        & 79.8 $\pm$ 1.5 & 65.9 $\pm$ 1.4 & 65.1 $\pm$ 6.1 & 60.6 $\pm$ 7.6 & 60.8 $\pm$ 10.1 & 86.3 $\pm$ 1.4 & 70.9 $\pm$ 1.4 & 58.7 $\pm$ 5.1 & 93.7 $\pm$ 6.4\\
        
        Digraph \cite{Tong2020DigraphIC}
        & 76.7 $\pm$ 1.9 & 62.9 $\pm$ 1.8 & 54.6 $\pm$ 6.8 & 76.7 $\pm$ 1.9 & 54.6 $\pm$ 6.8 & 83.9 $\pm$ 11.4 & 73.2 $\pm$ 11.7 & 49.1 $\pm$ 2.7 & 92.2 $\pm$ 14.1\\
        
        DiGCN \cite{Tong2020DigraphIC}
        & 76.5 $\pm$ 1.6 & 61.6 $\pm$ 1.9 & 54.3 $\pm$ 7.5 & 72.8 $\pm$ 7.7 & 65.6 $\pm$ 12.1 & 88.9 $\pm$ 0.6 & 83.4 $\pm$ 1.5 & 73.3 $\pm$ 15.3 & 97.4 $\pm$ 0.2 \\
        
        MagNet \cite{Zhang2021MagNetAN}
        &78.7 $\pm$ 2.2 & 64.6 $\pm$ 2.2 & 74.6 $\pm$ 4.4 & 77.1 $\pm$ 1.4 & 68.2 $\pm$ 7.0 & \textbf{89.8 $\pm$ 0.5} & 87.0 $\pm$ 0.6 & 78.6 $\pm$ 10.6 & 97.7 $\pm$ 0.2\\
        
        % $q$ & 0.00 & 0.05 & 0.15 & 0.15 & 0.15 & 0.15 & 0.15 & 0.25 & 0.25\\
        \hline
        
        \textbf{Framelet-MagNet} %Ours
        & \textbf{83.8 $\pm$1.4} & \textbf{67.8 $\pm$1.5} &  \textbf{77.0 $\pm$ 3.5} & \textbf{78.1 $\pm$ 1.2} & \textbf{73.8 $\pm$ 6.0} & 89.7 $\pm$ 0.4 & \textbf{88.5 $\pm$ 1.0} & \textbf{86.7 $\pm$ 5.7} & \textbf{97.8 $\pm$ 0.2}\\ 
        
        $\mathbf{q}$ & 0.00 &0.05 & 0.25 & 0.15 & 0.25 & 0.25 & 0.15 & 0.25 & 0.25\\
        
        \textbf{Framelet type} &  \texttt{Sigmoid} &  \texttt{Sigmoid} &  \texttt{Quadratic} &  \texttt{Haar} &  \texttt{Haar} &  \texttt{Linear} &  \texttt{Haar} &  \texttt{Linear} &  \texttt{Sigmoid}\\
        
        \hline
    \end{tabular}}
\end{table*}
We compare our model, Framelet-MagNet, with 10 state-of-the-art models in node classification, link prediction, and denoising. Link prediction consists of two different tasks, link existence prediction and link direction prediction. 

\subsection{Datasets and Implementation Details}
{}Node classification experiment is a semi-supervised task based on two citation datasets, CORA\_ML and CITESEER \cite{BojchevskiGuennemann2018}, and a WebKB dataset, CORNELL \cite{pei2020geom}. For citation datasets, following the experiments in \cite{kipf2017semisupervised}, we use 20 labels from each class for training, 500 labels for validation, and the rest for testing. For CORNELL, we use a 60\%/20\%/20\% train/validation/test split. Link prediction tasks are conducted on CORA\_ML, CORNELL, and a WikipediaNetwork dataset, CHAMELEON \cite{rozemberczki2021multi}. We remove 5\% edges for validation, 15\% edges for testing, and we keep the rest of the edges for training, such that the number of nodes remains constant after the split. The experiments are conducted on 10 random subsets from each dataset, and we will evaluate the models with the average results. We use node attributes as the input features for node classification. For link prediction, we identify edges through ordered node pairs and use in-degrees and out-degrees as input features to learn directly from the graph structure. Let $\mathcal{G}_d\{\mathcal{V}, \mathcal{E}, \mathbf{A}\}$ be a digraph. For an ordered node pair $v_i, v_j \in \mathcal{V}$, we define its label as following (1) existence prediction: 0 if $(v_i,v_j) \in \mathcal{E}$ and 1 otherwise; (2) direction prediction: 0 if $(v_i,v_j) \in \mathcal{E}$ and 1 if $(v_j,v_i) \in \mathcal{E}$. Note that we choose only linked node pairs for the link direction task. Moreover, we use either asymmetric or symmetrized adjacency matrix to train spatial models. ChebNet is trained with only symmetrized Laplacian. For the rest of the models, we adopt the asymmetric adjacency matrix and construct their special Laplacians accordingly. Hyperparameters including the number of filters, the learning rate, and magnetic parameter $q$ are tuned following grid search as common practice.

\begin{figure}[t]
\begin{minipage}[b]{1.0\linewidth}
  \centering
  \centerline{\includegraphics[height = 3cm, width=9cm]{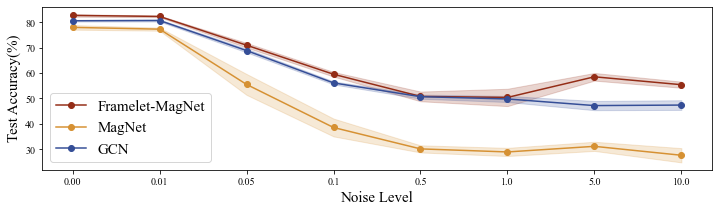}}
%  \vspace{2.0cm}
\end{minipage}
\caption{Denoising results on CORA\_ML. Framelet-MagNet (red) achieves higher classification accuracy at every noise level over other models.}
\label{fig:denoising}
\end{figure}

\subsection{Experiment Results}
The experiment results of node classification and link prediction tasks are shown in \textbf{Table \ref{Table_results}}. In node classification, Framelet-MagNet presents a good performance across all three datasets with the highest classification accuracy. It improves the state-of-the-art accuracy by 2.6\% on CORA\_ML and 2.4\% on CORNELL. Framelet-MagNet selects small $q$ for citation datasets. We suggest that this is because classification of articles does not depend on directional relationship. No matter one paper cites or is cited by another paper, they are likely to be in the same category. In link existence prediction, our model achieves the best accuracy on CORA\_ML and CORNELL, enhancing the state-of-the-art performance by 1\% and 5.6\%, respectively. In terms of CHAMELEON, GCN and MagNet have the highest accuracy of 89.8\% while our model achieves the second best with an accuracy of 89.7\%. In link direction prediction, our model again beats other approaches over all datasets. Especially, on CORNELL, Framelet-MagNet produces an accuracy that is 8.1\% higher than the second best model. Compared with the node classification experiment, the optimal $q$ value is larger, implying that directional information is very useful in link-level tasks.
% Generally speaking, our observation in the node classification and link prediction experiments demonstrates the power of framelet-based graph convolution in analyzing and processing digraph signals for predictions.

\subsection{Test of Denoising Capability}
In real-life applications, it is inevitable that some information such as adversarial examples is harmful to model prediction  \cite{zugner2018adversarial}. Such information is considered as ``noises''. Intuitively, removing noises from graph signals will enhance predictive performance. This procedure is known as ``denoising''. We test the denoising capability of Framelet-MagNet on the disturbed CORA\_ML dataset with node classification experiment settings. To perturb CORA\_ML whose features are normalized numbers, we manually impose Gaussian distributed noises and regulate noise level by altering the distribution standard deviation. \textbf{Fig.~\ref{fig:denoising}} presents the results, based on which we conclude that Framelet-MagNet shows clear superiority over MagNet and GCN in the denoising task. 

\section{Conclusion}
\label{sec:conclusion}
We propose Framelet-MagNet, a magnetic framelet-based spectral GCNN for digraphs, and demonstrate its power over state-of-the-art methods via empirical results. We realize framelet convolution and process digraph signals in a complex frequency domain to achieve effective filtering. However, due to the limitation of magnetic Laplacian, our method is not applicable to weighted mixed graphs. Besides, our link-level experiment has no clear definition of undirected edges, so we treat undirected information as noises. Future works may investigate the solutions to current limitations.

\textit{\textbf{Acknowledgements}} We would like to acknowledge Xitong Zhang for his suggestion on reproducing the experiment results in \cite{Zhang2021MagNetAN}.

% References should be produced using the bibtex program from suitable
% BiBTeX files (here: strings, refs, manuals). The IEEEbib.bst bibliography
% style file from IEEE produces unsorted bibliography list.
% -------------------------------------------------------------------------
\clearpage
%\textbf{Need to be shortened}
\bibliographystyle{IEEEbib}
\bibliography{refs}

\begin{thebibliography}{10}

\bibitem{zhang2019graph}
Si~Zhang, Hanghang Tong, Jiejun Xu, and Ross Maciejewski,
\newblock ``Graph convolutional networks: a comprehensive review,''
\newblock {\em Computational Social Networks}, vol. 6, no. 1, pp. 1--23, 2019.

\bibitem{xu2019graph}
Bingbing Xu, Huawei Shen, Qi~Cao, Yunqi Qiu, and Xueqi Cheng,
\newblock ``Graph wavelet neural network,''
\newblock {\em arXiv:1904.07785}, 2019.

\bibitem{zheng2021howframelet}
Xuebin Zheng, Bingxin Zhou, Junbin Gao, Yuguang Wang, Pietro Li{\'{o}}, Ming
  Li, and Guido Mont{\'{u}}far,
\newblock ``How framelets enhance graph neural networks,''
\newblock in {\em ICML}, 2021.

\bibitem{yang2022quasi}
Mengxi Yang, Xuebin Zheng, Jie Yin, and Junbin Gao,
\newblock ``Quasi-framelets: Another improvement to graph neural networks,''
\newblock {\em arXiv:2201.04728}, 2022.

\bibitem{an2004characterizing}
Yuan An, Jeannette Janssen, and Evangelos~E Milios,
\newblock ``Characterizing and mining the citation graph of the computer
  science literature,''
\newblock {\em Knowledge and Information Systems}, vol. 6, no. 6, pp. 664--678,
  2004.

\bibitem{abedin2009graph}
Babak Abedin and Babak Sohrabi,
\newblock ``Graph theory application and web page ranking for website link
  structure improvement,''
\newblock {\em Behaviour \& Information Technology}, vol. 28, no. 1, pp.
  63--72, 2009.

\bibitem{li2018diffusion}
Yaguang Li, Rose Yu, Cyrus Shahabi, and Yan Liu,
\newblock ``Diffusion convolutional recurrent neural network: Data-driven
  traffic forecasting,''
\newblock in {\em ICLR}, 2018.

\bibitem{monti2018motifnet}
Federico Monti, Karl Otness, and Michael~M Bronstein,
\newblock ``Motifnet: a motif-based graph convolutional network for directed
  graphs,''
\newblock in {\em IEEE DSW}, 2018, pp. 225--228.

\bibitem{ma2019spectral}
Yi~Ma, Jianye Hao, Yaodong Yang, Han Li, Junqi Jin, and Guangyong Chen,
\newblock ``Spectral-based graph convolutional network for directed graphs,''
\newblock {\em arXiv:1907.08990}, 2019.

\bibitem{Tong2020DigraphIC}
Zekun Tong, Yuxuan Liang, Changsheng Sun, Xinke Li, David~S. Rosenblum, and
  Andrew Lim,
\newblock ``Digraph inception convolutional networks,''
\newblock in {\em NeurIPS}, 2020.

\bibitem{Zhang2021MagNetAN}
Xitong Zhang, Yixuan He, Nathan Brugnone, Michael Perlmutter, and Matthew~J.
  Hirn,
\newblock ``{MagNet}: A neural network for directed graphs,''
\newblock in {\em NeurIPS}, 2021.

\bibitem{fanuel2017magnetic}
Micha{\"e}l Fanuel, Carlos~M Alaiz, and Johan~AK Suykens,
\newblock ``Magnetic eigenmaps for community detection in directed networks,''
\newblock {\em Physical Review E}, vol. 95, no. 2, pp. 022302, 2017.

\bibitem{zou2022simple}
Chunya Zou, Andi Han, Lequan Lin, and Junbin Gao,
\newblock ``A simple yet effective {SVD-GCN} for directed graphs,''
\newblock {\em arXiv:2205.09335}, 2022.

\bibitem{dong2017sparse}
Bin Dong,
\newblock ``Sparse representation on graphs by tight wavelet frames and
  applications,''
\newblock {\em Applied and Computational Harmonic Analysis}, vol. 42, pp.
  452--479, 2017.

\bibitem{zheng2022decimated}
Xuebin Zheng, Bingxin Zhou, Yu~Guang Wang, and Xiaosheng Zhuang,
\newblock ``Decimated framelet system on graphs and fast g-framelet
  transforms.,''
\newblock {\em Journal of Machine Learning Research}, vol. 23, pp. 18--1, 2022.

\bibitem{hammond2011wavelets}
David~K Hammond, Pierre Vandergheynst, and R{\'e}mi Gribonval,
\newblock ``Wavelets on graphs via spectral graph theory,''
\newblock {\em Applied and Computational Harmonic Analysis}, vol. 30, no. 2,
  pp. 129--150, 2011.

\bibitem{defferrard2016convolutional}
Micha{\"e}l Defferrard, Xavier Bresson, and Pierre Vandergheynst,
\newblock ``Convolutional neural networks on graphs with fast localized
  spectral filtering,''
\newblock in {\em NeurIPS}, 2016.

\bibitem{kipf2017semisupervised}
Thomas~N. Kipf and Max Welling,
\newblock ``Semi-supervised classification with graph convolutional networks,''
\newblock in {\em ICLR}, 2017.

\bibitem{klicpera2019predict}
Johannes Klicpera, Aleksandar Bojchevski, and Stephan G{\"{u}}nnemann,
\newblock ``Predict then propagate: Graph neural networks meet personalized
  pagerank,''
\newblock in {\em ICLR}, 2019.

\bibitem{hamilton2017inductive}
Will Hamilton, Zhitao Ying, and Jure Leskovec,
\newblock ``Inductive representation learning on large graphs,''
\newblock in {\em NeurIPS}, 2019.

\bibitem{xu2019how}
Keyulu Xu, Weihua Hu, Jure Leskovec, and Stefanie Jegelka,
\newblock ``How powerful are graph neural networks?,''
\newblock in {\em ICLR}, 2019.

\bibitem{velivckovic2017graph}
Petar Veli{\v{c}}kovi{\'c}, Guillem Cucurull, Arantxa Casanova, Adriana Romero,
  Pietro Lio, and Yoshua Bengio,
\newblock ``Graph attention networks,''
\newblock in {\em ICLR}, 2018.

\bibitem{tong2020directed}
Zekun Tong, Yuxuan Liang, Changsheng Sun, David~S Rosenblum, and Andrew Lim,
\newblock ``Directed graph convolutional network,''
\newblock {\em arXiv:2004.13970}, 2020.

\bibitem{BojchevskiGuennemann2018}
Aleksandar Bojchevski and Stephan G\"{u}nnemann,
\newblock ``Deep {Gaussian} embedding of graphs: Unsupervised inductive
  learning via ranking,''
\newblock in {\em ICLR}, 2018.

\bibitem{pei2020geom}
Hongbin Pei, Bingzhe Wei, Kevin~Chen{-}Chuan Chang, Yu~Lei, and Bo~Yang,
\newblock ``{Geom-GCN}: Geometric graph convolutional networks,''
\newblock in {\em ICLR}, 2020.

\bibitem{rozemberczki2021multi}
Benedek Rozemberczki, Carl Allen, and Rik Sarkar,
\newblock ``Multi-scale attributed node embedding,''
\newblock {\em Journal of Complex Networks}, vol. 9, no. 2, pp. cnab014, 2021.

\bibitem{zugner2018adversarial}
Daniel Z{\"u}gner, Amir Akbarnejad, and Stephan G{\"u}nnemann,
\newblock ``Adversarial attacks on neural networks for graph data,''
\newblock in {\em ACM SIGKDD}, 2018, pp. 2847--2856.

\end{thebibliography}

\end{document}